# DFE-IANet: A Method for Polyp Image Classification Based on Dual-domain Feature Extraction and Interaction Attention

Wei Wang, Jixing He and Xin Wang[✉]

School of Computer and Communication Engineering, Changsha University of Science and Technology, Changsha 410114, China
`wangxin@csust.edu.cn`

**Abstract.** It is helpful in preventing colorectal cancer to detect and treat polyps in the gastrointestinal tract early. However, there have been few studies to date on designing polyp image classification networks that balance efficiency and accuracy. This challenge is mainly attributed to the fact that polyps are similar to other pathologies and have complex features influenced by texture, color, and morphology. In this paper, we propose a novel network DFE-IANet based on both spectral transformation and feature interaction. Firstly, to extract detailed features and multi-scale features, the features are transformed by the multi-scale frequency domain feature extraction (MSFD) block to extract texture details at the fine-grained level in the frequency domain. Secondly, the multi-scale interaction attention (MSIA) block is designed to enhance the network's capability of extracting critical features. This block introduces multi-scale features into self-attention, aiming to adaptively guide the network to concentrate on vital regions. Finally, with a compact parameter of only 4M, DFE-IANet outperforms the latest and classical networks in terms of efficiency. Furthermore, DFE-IANet achieves state-of-the-art (SOTA) results on the challenging Kvasir dataset, demonstrating a remarkable Top-1 accuracy of 93.94%. This outstanding accuracy surpasses ViT by 8.94%, ResNet50 by 1.69%, and VMamba by 1.88%. Our code is publicly available at `https://github.com/PURSUETHESUN/DFE-IANet`.

## 1 Introduction

Colorectal cancer (CRC) is the third most prevalent cancer worldwide [1]. It is primarily caused by the malignant transformation of normal cells in the colon or rectum, and this malignant transformation frequently presents as precancerous polyps. Early detection and removal of polyps are beneficial in preventing CRC. In clinical diagnosis, clinicians frequently have to manually identify images from colonoscopy images. However, this process requires a significant amount of medical resources. Furthermore, the classification's accuracy and efficiency may vary greatly due to differences in the diagnostic abilities of different clinicians. Hence, there is an immediate requirement for a precise and efficient method to automatically classify polyp images. It is critical for clinicians to make further diagnoses and assessments of lesions based on the classification results.

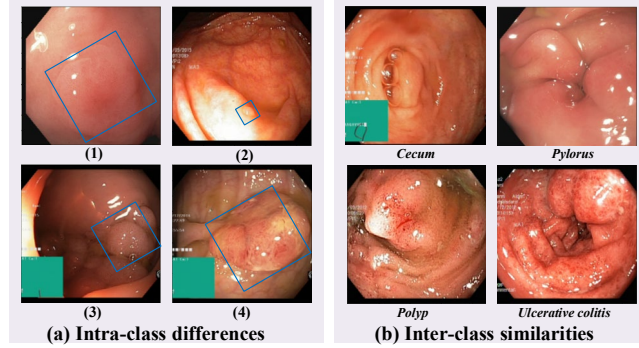

**Fig. 1.** There are some samples from the Kvasir dataset [2].

Gastrointestinal diseases and tissues exhibit extremely complex characteristics. As shown in Fig. 1(a), the size and morphology of polyps show significant variations at different periods and locations. In the early stages, polyps are relatively small and do not show significant boundaries, which may lead to missed detection of polyps. In the later stages, polyps again present different sizes and textures, which may be incorrectly identified as other tissues. In Fig. 1(b), polyps share similar structures and characteristics with the other three common pathologies and tissues, such as elevated and textural features. Furthermore, colonoscopy images can be affected by various factors like blood, mucus, and brightness, resulting in blurred images, noise, and interfering objects. These factors can impact the accuracy of polyp image classification.

Currently, Convolutional Neural Network (CNN) and Transformer-based [3] classification models have gained significant popularity. Patel et al. [4] evaluated the accuracy of polyp classification by using general networks, demonstrating a noteworthy improvement obtained by CNN-based research. Wu et al. [5] employed pre-trained ViT [6] in the CT image classification of emphysema to overcome insufficient data, attaining greater accuracy than CNN. The utilization of ViT by Komorowski et al. [7] aimed at enhancing the classification accuracy of chest X-ray images. Liu et al. [8] proposed a new feature pyramid visual Transformer that combines the local features with the global features, leading to a breakthrough in accuracy. However, there are still certain limitations that need to be acknowledged. On the one hand, a simple combination of CNN and Transformer may not be able to achieve a high accuracy to meet practical needs. On the other hand, using pre-trained high-performance networks still consumes expensive hardware resources. Therefore, one of the major challenges at present is how to effectively balance accuracy and efficiency to design a specific network that is targeted to solve the difficult problems in polyp image classification.

To settle the challenges above, we propose a novel network DFE-IANet for polyp image classification that leverages spectral transformation and interaction attention. First, to capture texture details of polyps, the spatial features undergo the discrete wavelet transform (DWT). Then, these texture details are extracted using asymmetric depth-wise (ADW) convolution on different detail components. To the best of our knowledge, this is the first time that ADW is used to extract fine-grained features based on different features in the frequency domain. Second, to enhance the network's capability to perceive critical features, a new multi-scale interaction attention is proposed. This approach enhances the network's capability to capture relevant and discriminative information. Finally, considering the distribution of texture details and semantic features in the network, a multi-branch multi-scale feature extraction layer is introduced in the shallow layers, while a cascade multi-scale feature extraction layer is incorporated in the deep layers. Our work makes the following key contributions:

1. To extract texture details, the features undergo a spatial-to-frequency domain transformation, which extracts detailed features of polyps on different detail components. Moreover, the multi-branch multi-scale feature extraction layer is designed to capture features at various scales. As a result, the multi-scale frequency domain feature extraction (MSFD) block is designed to efficiently capture detailed features in the shallow layer.
2. An adaptive feature guidance layer based on multi-scale interaction attention is proposed. By modeling the dependencies among multi-scale features, it aims to guide the network to extract critical features adaptively. Furthermore, a cascade multi-scale feature extraction layer is devised to extract rich semantic information. Thus, the multi-scale interaction attention (MSIA) block is constructed to extract features in the deeper layers.
3. A novel neural network DFE-IANet based on MSFD and MSIA blocks is proposed for polyp image classification. Through comparative experiments with both classical and newest networks, the proposed DFE-IANet achieves state-of-the-art accuracy while also prioritizing efficiency.

## 2 Related work

### 2.1 CNN and Transformer for medical image classification

Recently, there has been a notable surge in the popularity of CNN and Transformer-based image classification methods. To tackle the challenge of colon lesion classification, DLGNet [9] proposed a dual-branch lesion-aware method that achieved 91.50% accuracy on the colon lesion dataset. Wei et al. [10] verified that networks based on deep learning outperform clinicians in accuracy by conducting massive experiments on multiple datasets. O-Net [11] is a two-branch network that combines Transformer and CNN to achieve high accuracy. Huo et al. [12] designed a feature fusion network by fusing global and local features, which achieved high accuracies in polyp image classification. However, traditional networks still have high computational complexity simply combining local and global feature extraction, which limits their application.

### 2.2    Feature extraction methods in the frequency domain

Images can undergo spectral transformation into various frequency components within the frequency domain. Rippel et al. [13] demonstrated that converting images into different frequency components offers both powerful feature representations. Ding et al. [14] combined convolution and wavelet transform to enhance feature representation for improving performance in the classification of ultrasound images. To tackle the loss of information in the spatial domain, Xu et al. [15] employed the discrete cosine transform to convert features. Besides, Huang et al. [16] proposed an adaptive frequency filter, which achieves the effect of large convolution kernels through elemental multiplication in the frequency domain. Different from the above methods, we employ ADW convolution to extract texture details in the frequency domain.

### 2.3    Effective feature interactions

Spatial feature interactions are efficient for networks to perceive contextual information. ViT [6] employed self-attention mechanisms to facilitate global feature interactions. This approach has proven to be highly effective and has demonstrated superior performance in various vision tasks. Building upon the concept of self-attention, Yang et al. [17] proposed a focal modulation module to enhance feature interactions. This module demonstrated superior performance compared to ViT. Additionally, Rao et al. [18] introduced a recursive gated convolution that leveraged high-order feature interactions, improving performance in respective tasks. Li et al. [19] introduced the concept of multi-order gated aggregation networks, which demonstrated strong performance across various tasks, including image classification. The above studies show that multi-scale feature interactions during extracting features will help to raise the capability of feature extraction. In contrast to the mentioned studies, we propose a feature guidance layer based on multi-scale feature interactions to adaptively guide the network's attention towards important region features.

## 3    Methodology

### 3.1    Macro architecture of DFE-IANet

The macro architecture of DFE-IANet is depicted in Fig. 2, showcasing its detailed components. In the deep neural network, the shallow layers contain more texture details, whereas the deep layers have richer semantic features. Since texture details involve crucial features of polyps, it is critical to capture texture details in the shallow layers. Therefore, fine-grained features are extracted in shallow layers using the MSFD block. As the network becomes progressively deeper, the semantic information it contains is richer. The MSIA block is employed to model the dependencies among multi-scale features. It serves the purpose of adaptively guiding the network to focus on the crucial features of polyps. Finally, the specific number of blocks is [2, 3, 5, 2] in each stage [20].

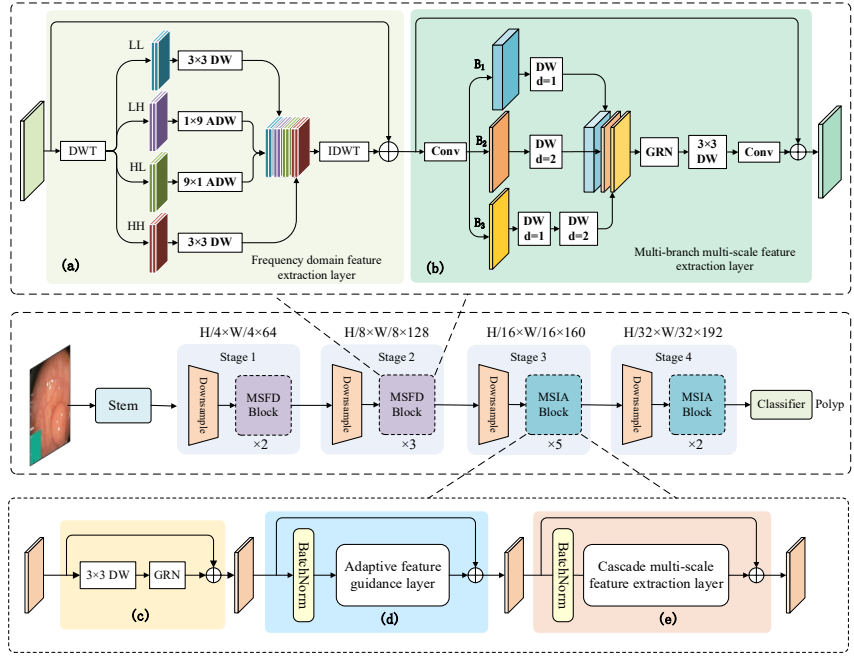

**Fig. 2.** Macro architecture of DFE-IANet: modules (a) ∼ (b) make up the MSFD block, and modules (c) ∼ (e) make up the MSIA block.

### 3.2 Multi-scale frequency domain feature extraction (MSFD)

**Frequency domain feature extraction layer.** The texture details contain numerous key features of polyps, and these detailed features are essential for accurately identifying polyps. Therefore, the frequency domain feature extraction layer (Fig. 2(a)) is designed, which utilizes DWT to transform the features into the frequency domain [21]. To be more specific, DWT is applied to the input feature x ∈ $\mathbb{R}^{C \times H \times W}$ to obtain four different detail components: the approximate component $X_{LL} \in \mathbb{R}^{C \times H/2 \times W/2}$, the horizontal detail component $X_{LH} \in \mathbb{R}^{C \times H/2 \times W/2}$, the vertical detail component $X_{HL} \in \mathbb{R}^{C \times H/2 \times W/2}$ and the diagonal detail component $X_{HH} \in \mathbb{R}^{C \times H/2 \times W/2}$. Then, common 3×3 DW convolutions are used to extract detailed features on the $X_{LL}$ and $X_{HH}$. The detailed features on the $X_{LH}$ and $X_{HL}$ are extracted by applying 1×9 and 9×1 ADW convolutions, respectively. The output features are reconstructed using the inverse discrete wavelet transform (IDWT).

$$[X_{LL}, X_{LH}, X_{HL}, X_{HH}] = DWT(x) \qquad (1)$$

$$\hat{X} = concat(DW_{LL}(X_{LL}), ADW_{LH}(X_{LH}), ADW_{HL}(X_{HL}), DW_{HH}(X_{HH})) \qquad (2)$$

$$Z = IDWT(\hat{X}) + x \qquad (3)$$

6        W. Wang et al.

where $DW_{LL}$ and $DW_{HH}$ are 3×3 depth-wise convolution operations. $ADW_{LH}$ and $ADW_{HL}$ are 1×9 and 9×1 asymmetric depth-wise convolution operations, respectively. *concat* is the concatenation operation along the channel dimension.

**Multi-branch multi-scale feature extraction layer.** The specific structure of the multi-branch multi-scale feature extraction layer can be seen in Fig. 2(b). In response to the classification of images from colonoscopy, small-scale pathological features account for the majority, such as polyps, ulcerative colitis, esophagitis, etc. As a result, more input features will be assigned to the branch that extracts small-scale features. Specifically, the features are split along the channel in the 2:1:1 ratio to produce three distinct groups of features: $z_1 \in \mathbb{R}^{2C \times H \times W}$, $z_2 \in \mathbb{R}^{C \times H \times W}$, $z_3 \in \mathbb{R}^{C \times H \times W}$. The obtained features are then fed into separate branches to capture multi-scale information. Subsequently, a global response normalization (GRN) [22] operation is applied to normalize the features. Subsequently, these features are processed using 3×3 DW convolution, which avoids the gridding effect [23]. Lastly, the features are fused by a convolution layer.

$$[z_1, z_2, z_3] = Split(Conv_{1 \times 1}(Z)) \qquad (4)$$

$$z = DW_{d=1}(GRN(GELU(concat(B_1(z_1), B_2(z_2), B_3(z_3))))) \qquad (5)$$

$$Z' = Conv_{1 \times 1}(z) + Z \qquad (6)$$

where $B_1$, $B_2$, and $B_3$ are three different branches shown in Fig. 2(b). *Split* is a splitting operation along the channel dimension. *concat* refers to the concatenation along the channel.

### 3.3  Multi-scale interaction attention (MSIA)

**Conditional positional encoding layer.** The position information is crucial in classification methods based on Transformer, which directly affects the overall performance of the network. In previous research, mostly absolute positional encoding or relative positional encoding was used. Recently, Chu et al. [24] pointed out that conditional positional encoding (CPE) achieves better performance through simple operations such as convolution. To embed conditional position information, the 3×3 DW convolution is employed within the MSIA block. The structure of the CPE can be visualized in Fig. 2(c). The process is represented mathematically by the following equation:

$$x' = GRN(DW(x)) + x \qquad (7)$$

where $x' \in \mathbb{R}^{C \times H \times W}$ is the output feature embedded CPE, and DW is the CPE operation through using 3×3 depth-wise convolution.

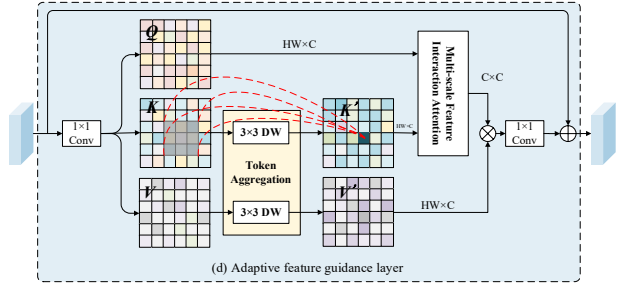

**Fig. 3.** Adaptive feature guidance layer in Fig. 2(d).

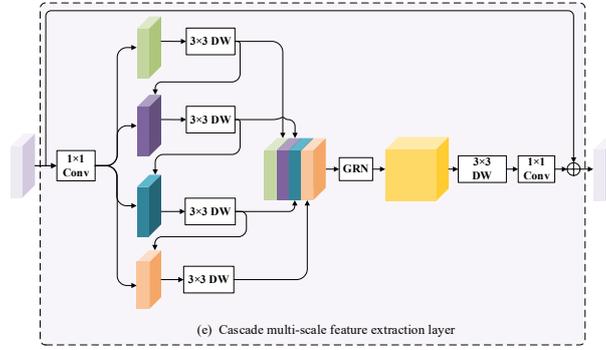

**Fig. 4.** Cascade multi-scale feature extraction layer in Fig. 2(e).

**Adaptive feature guidance layer.** Capturing key discriminative features is crucial. However, traditional self-attention models lack the incorporation of multi-scale feature interactions. To address this issue, integrating multi-scale features into self-attention enables the network to dynamically guide its focus towards important regions (Fig. 3). Specifically, before $Q$ interacts with $K$, the features of neighboring tokens in $K$ and $V$ are firstly aggregated using 3×3 DW convolution to obtain $K'$ and $V' \in \mathbb{R}^{HW \times C}$, respectively. Thus, each token in $K'$ and $V' \in \mathbb{R}^{HW \times C}$ aggregates the features of neighboring tokens, which differ in terms of scale from the tokens in $Q$. Next, $Q$ and $K'$ are used to compute the attention score, which implements the muti-scale feature interactions between the token and local features. The attention score is then multiplied with $V'$ after $Softmax$. The final output features are obtained by applying a 1×1 convolution layer.

$$[Q, K, V] = Split(Conv_{1 \times 1}(x')) \tag{8}$$

$$K' = DW_{3 \times 3}(K) \tag{9}$$

$$V' = DW_{3 \times 3}(V) \tag{10}$$

$$AttenMat(Q, K', V') = Softmax(\frac{QK'^T}{\sqrt{d}})V' \qquad (11)$$

$$Z = Conv_{1\times1}(AttenMat(Q, K', V')) + x' \qquad (12)$$

where $K'$ and $V'$ are the features after aggregating neighboring tokens. *Split* is the slicing operation along the channel dimension, and $DW_{3\times3}$ is aggregation operation using 3×3 DW convolution.

**Cascade multi-scale feature extraction layer**. Since the deeper layers of deep neural networks contain more semantic features, extracting rich semantic features is crucial for improving classification accuracy. In addition, input features in the feed-forward neural network (FFN) layer of ViT are only fused among channels, while there is a lack of spatial feature fusion. Therefore, the cascade multi-scale feature extraction layer (Fig. 4) is designed. Specifically, the input features are split along the channel dimension by a 1:1:1:1 rate. The different features are sequentially extracted by cascade DW convolution to obtain four groups of semantic features. These features are concatenated along the channel. To enhance local features, 3×3 DW convolutions are utilized for feature extraction. Subsequently, using a 1×1 convolution layer fuses different features.

$$[z_1, z_2, z_3, z_4] = Split(Conv_{1\times1}(Z)) \qquad (13)$$

$$z'_1 = DW_{3\times3}(z_1) \qquad (14)$$

$$z'_2 = DW_{3\times3}(z_2 + z'_1) \qquad (15)$$

$$z'_3 = DW_{3\times3}(z_3 + z'_2) \qquad (16)$$

$$z'_4 = DW_{3\times3}(z_4 + z'_3) \qquad (17)$$

$$Z' = Conv_{1\times1}(DW_{3\times3}(GRN(GELU(concat(z'_1, z'_2, z'_3, z'_4))))) + Z \qquad (18)$$

where $z_1, z_2, z_3, z_4 \in \mathbb{R}^{C\times H\times W}$ are the different features after splitting along channel. *Split* is the channel slicing operation, and *concat* is the operation of concatenating along the channel.

## 4      Experiments

### 4.1      Datasets

To assess the performance of the proposed network, we conducted experiments on two datasets: the Kvasir dataset [2] and the colonic polyp (CP) dataset [25], and the Kvasir dataset is used for ablation experiments. Both datasets exhibit significant intra-class variance and high similarity between different pathologies.

The Kvasir dataset is a public dataset with eight classes. These classes contain polyps, ulcerative colitis, esophagitis, etc. Each class contains 1000 images, with a total of 8000 images. The images in the Kvasir dataset have resolutions ranging from 720×576 to 1920×1072 pixels. In the experiments, the images from each class are divided into a training set and a test set in a 4:1 ratio to ensure a fair evaluation. Specifically, 6400 images are allocated to the training set, while 1600 images are assigned to the test set.

The CP dataset is a colonoscopy dataset obtained from a grade A tertiary hospital, which has been utilized in prior studies to investigate colonic polyp classification. The dataset comprises a collection of medical images acquired from actual patients at the hospital, ensuring its authenticity and relevance to real-world scenarios. The dataset has undergone meticulous curation and annotation processes, providing accurate and reliable labels for each image. In this dataset, the specific number of images in each class is 2199 for polyps, 743 for ulcerative colitis, 5500 for normal, and 2167 for other lesions, respectively. The CP dataset consists of a total of 10609 images. Each image is cropped to a fixed resolution of 256×256 pixels. To conduct the experiments, the dataset is divided into training and test sets in a 4:1 ratio. This results in 8,489 images allocated to the training set and 2120 images assigned to the test set.

### 4.2 Implementation details

All experiments are conducted using Python 3.9.12 with PyTorch 1.10.0 framework, and the training is performed on an A10 GPU with CUDA 11.2 support. Data enhancement strategies are used to get sufficient data samples. The image preprocessing begins by cropping the images to a fixed 256×256. Subsequently, a center cropping operation is performed to further resize them to 224×224. Additionally, random horizontal flipping and normalization techniques are used to augment the data during training. The initial learning rate is set to 0.0005, weight decay is set to 0.05, and the Adamw optimizer [26] is utilized. A batch size of 16 is used, and the training is performed for 400 epochs. The cross-entropy loss is used to train the models. We utilize commonly used evaluation metrics in image classification, including Top-1 Accuracy (Acc), Precision (Pre), Recall (Rec), Specificity (Spe), and F1-score (F1). The formula for each evaluation criterion is shown below:

$$Accuracy = \frac{T_P + T_N}{T_P + T_N + F_P + F_N} \tag{19}$$

$$Precision = \frac{T_P}{T_P + F_P} \tag{20}$$

$$Recall = \frac{T_P}{T_P + F_N} \tag{21}$$

$$Specifity = \frac{T_N}{T_N + F_P} \tag{22}$$

$$F1 - score = \frac{2 \times T_P}{2 \times T_P + F_P + F_N} \tag{23}$$

where $T_P$ represents True Positive, $F_N$ represents False Negative, $F_P$ represents False Positive, and $T_N$ represents True Negative.

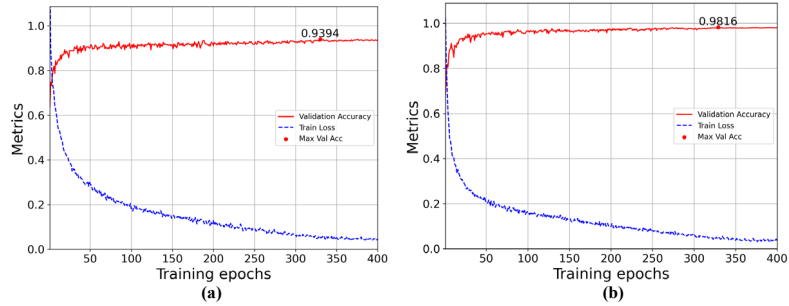

**Fig. 5.** Training metrics curves of the proposed DFE-IANet on the Kvasir dataset (a) and CP dataset (b), respectively.

**Table 1.** Experimental results of different networks on the Kvasir dataset (%).

| Model | Params (M) | Flops (G) | Acc | Pre | Rec | Spe | F1 |
|---|---|---|---|---|---|---|---|
| ResNet50 [27] | 23.52 | 4.13 | 92.25 | 92.34 | 92.25 | 98.89 | 92.25 |
| FocalNet [17] | 27.66 | 4.42 | 90.69 | 90.79 | 90.69 | 98.67 | 90.70 |
| ViT [6] | 85.65 | 16.8 | 85.00 | 85.17 | 85.00 | 97.86 | 84.96 |
| Swin [28] | 27.50 | 4.37 | 90.62 | 90.64 | 90.62 | 98.66 | 90.63 |
| CMT [29] | 13.82 | 2.07 | 91.94 | 91.95 | 91.94 | 98.85 | 91.93 |
| Wave-Vit [21] | 21.80 | 4.37 | 91.44 | 91.46 | 91.44 | 98.78 | 91.45 |
| InternImage [30] | 29.16 | 4.82 | 90.62 | 90.67 | 90.62 | 98.66 | 90.62 |
| CoCs [31] | 14.13 | 2.77 | 91.56 | 91.63 | 91.56 | 98.79 | 91.56 |
| InceptionNeXt [32] | 25.76 | 4.20 | 91.63 | 91.66 | 91.63 | 98.80 | 91.61 |
| BiFormer [33] | 12.63 | 2.09 | 91.37 | 91.43 | 91.37 | 98.77 | 91.38 |
| HiFuse [12] | 118.05 | 17.44 | 90.94 | 91.05 | 90.94 | 98.71 | 90.95 |
| VMamba [34] | 38.13 | 6.32 | 92.06 | 92.10 | 92.06 | 98.87 | 92.07 |
| **DFE-IANet (ours)** | **4.00** | **1.21** | **93.94** | **93.96** | **93.94** | **99.13** | **93.94** |

### 4.3    Comparison with the state-of-the-art networks

In this section, 12 representative networks are chosen for comparison experiments. All experiments are conducted from scratch in a consistent environment, and the evaluation metric is the Top-1 accuracy. As depicted in Table 1, the proposed DFE-IANet is less than some other networks in terms of parameter count and computational cost. This demonstrates the superiority of DFE-IANet in achieving high performance with relatively efficient resource utilization. Fig. 5 displays the validation accuracy and training loss curves during the training of DFE-IANet on both the Kvasir and CP datasets. The proposed DFE-IANet ultimately shows high accuracy and excellent convergence on both datasets.

Table 2. Experimental results of different networks on the CP dataset (%).

| Model | Year | Acc | Pre | Rec | Spe | F1 |
|---|---|---|---|---|---|---|
| ResNet50 [27] | 2016 | 97.55 | 95.64 | 94.91 | 99.18 | 95.26 |
| FocalNet [17] | 2022 | 97.08 | 94.84 | 93.62 | 99.04 | 94.18 |
| ViT [6] | 2020 | 95.14 | 93.09 | 90.99 | 98.10 | 91.98 |
| Swin [28] | 2021 | 96.93 | 95.54 | 94.46 | 98.80 | 94.98 |
| CMT [29] | 2022 | 97.88 | 96.03 | 94.92 | 99.30 | 95.44 |
| Wave-Vit [21] | 2022 | 97.64 | 95.01 | 95.22 | 99.26 | 95.11 |
| InternImage [30] | 2023 | 96.56 | 93.51 | 93.47 | 98.84 | 93.49 |
| CoCs [31] | 2023 | 97.08 | 94.76 | 93.81 | 99.05 | 94.24 |
| InceptionNeXt [32] | 2023 | 97.74 | 95.98 | 94.26 | 99.27 | 95.03 |
| BiFormer [33] | 2023 | 97.64 | 95.50 | 94.92 | 99.23 | 95.20 |
| HiFuse [12] | 2024 | 97.36 | 95.51 | 94.01 | 99.13 | 94.69 |
| VMamba [34] | 2024 | 97.74 | 96.36 | 94.56 | 99.23 | 95.38 |
| **DFE-IANet (ours)** | 2024 | **98.16** | **96.72** | **95.41** | **99.42** | **96.01** |

In Table 1, DFE-IANet attains an accuracy of 93.94% on the Kvasir dataset. Compared to ResNet50, DFE-IANet is 1.69% higher, outperforming representative CNN-based networks. Due to its inability to extract global features, the networks represented by CNN show disadvantages. The accuracy of DFE-IANet is higher than ViT and Swin-Transformer by 8.94% and 3.32%, respectively, which confirms that ViT and Swin-Transformer are suitable for cases with sufficient data samples. Although the latest networks (e.g., InternImage, BiFormer, and InceptionNext) all have higher classification accuracies than the traditional ones, DFE-IANet achieves more than a 2% improvement. This suggests that state-of-the-art classification networks cannot be targeted to address the challenges in polyp image classification. Table 2 displays the results of experiments on the CP dataset. HiFuse, the latest network for medical image classification, fails to perform superiorly on both of the datasets. In particular, DFE-IANet has less computational cost and fewer parameter numbers than HiFuse. In addition, the accuracy of VMamba on both datasets is still lower than that of DFE-IANet, which may be due to the fact that its scanning mechanism still has some limitations in extracting the local features. Overall, the proposed DFE-IANet achieves excellent performance on these evaluation metrics.

### 4.4  Ablation experiments

**Ablation experiments of asymmetric depth-wise convolution.** At the frequency domain feature extraction layer of Fig. 2(a), two additional groups of ADW convolutions are chosen to conduct ablation experiments. Table 3 shows that DFE-IANet has the least loss of performance using 1×7 and 7×1 ADW convolutions. However, the performance loss is relatively large when 1×11 and 11×1 ADW convolutions are used. This may be because 1×9 and 9×1 convolution kernels are more easily to capture fine-grained features.

Table 3. Ablation experiment results of asymmetric depth-wise convolution.

| Model | Params(M) | Flops(M) | Accuracy(%) | Kernel size |
|---|---|---|---|---|
| DFE-IANet (7) | **3.99** | **1210.02** | 93.87 | 1×7, 7×1 |
| DFE-IANe (11) | 4.00 | 1211.42 | 92.94 | 1×11, 11×1 |
| DFE-IANet (baseline) | 4.00 | 1210.72 | **93.94** | 1×9, 9×1 |

**Experiments of multi-branch multi-scale feature extraction layer.** To compare with larger convolutional kernels, ablation experiments are performed at the multi-branch multi-scale feature extraction layer in Fig. 2(b). To be more specific, the second branch is substituted with 5×5 DW convolution. The third branch is replaced with 7×7 DW convolution. Table 4 shows that large convolutional kernels lead to a reduction in the accuracy by 0.69%. It may be related to the fact that large convolutional kernels are to be trained with more difficulty.

Table 4. Ablation experiment results of multi-branch multi-scale feature extraction.

| Model | Params(M) | Flops(M) | Accuracy(%) | Convolutions |
|---|---|---|---|---|
| DFE-IANet (large) | 4.02 | 1243.73 | 93.25 | 5×5, 7×7 DW |
| DFE-IANet | **4.00** | **1210.72** | **93.94** | ours |

**Experiments of adaptive feature guidance layer.** The adaptive feature guidance layer (Fig. 3) exhibits better performance during feature extraction. Compared with using traditional self-attention (TS), ablation experiments are conducted to further validate the effectiveness of multi-scale feature interactions. The results of experiments in Table 5 display that incorporating TS into DFE-IANet leads to a decrease in accuracy by 0.75%. This indicates that the capability of extracting features can be enhanced through multi-scale feature interactions.

Table 5. Ablation experiment results of adaptive feature guidance layer.

| Model | Params(M) | Flops(M) | Accuracy(%) | Description |
|---|---|---|---|---|
| DFE-IANet (TS) | **3.98** | **1208.60** | 93.19 | Self-attention |
| DFE-IANet | 4.00 | 1210.72 | **93.94** | Feature interactions |

**Ablation experiments of MSFD and MSIA blocks**. In Table 6, the classification accuracy is 92.94% when utilizing only the MSFD block, whereas using only the MSIA block is an accuracy of 93.25%. However, it is worth stating that the high computational complexity of self-attention, greatly increases the memory footprint during training when only MSIA blocks are used. This observation sheds light on why certain existing networks use CNN in the shallow layers and incorporating Transformer-based self-attention in the deeper layers. This hybrid architecture aims to strike a balance between computational efficiency and capturing long-range dependencies in the feature extraction process.

**Table 6.** The Ablation experiments separately using MSFD and MSIA blocks.

| MSFD | MSIA | Accuracy(%) | Params(M) | Flops(M) |
|---|---|---|---|---|
| √ | × | 92.94 | 5.02 | **1113.92** |
| × | √ | 93.25 | **3.69** | 1931.92 |
| √ | √ | **93.94** | 4.00 | 1210.72 |

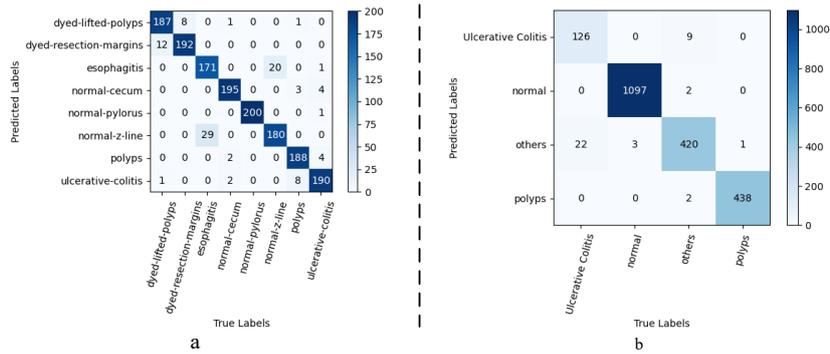

**Fig. 6.** The confusion matrix on the Kvasir dataset (a) and CP dataset (b), separately.

### 4.5   Visualization and analysis

**Confusion matrix analysis**. The confusion matrix can clearly show the results of the classification of each class. Fig. 6(a) displays the confusion matrix on the Kvasir dataset. Polyps are easily misidentified as ulcerative colitis, mainly because of the presence of disturbances such as blood stains on the surface of the polyps. In particular, distinguishing between esophagitis and normal z-line can be challenging due to that the esophagitis manifests as mucosal damage along the z-line. Besides, Fig. 6(b) illustrates the confusion matrix obtained on the CP dataset. The results indicate that DFE-IANet achieves higher sensitivity in classifying polyps, normal samples, and other classes. However, the accuracy of ulcerative colitis is relatively low, which may be related to insufficient samples.

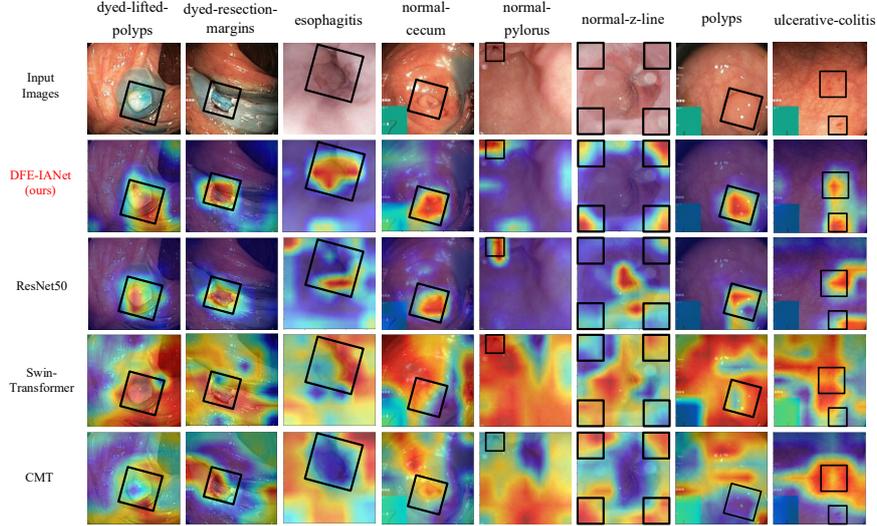

**Fig. 7.** The heat map results on the Kvasir dataset are presented. The first row showcases the input images, while the subsequent rows display the heat maps generated by DFE-IANet and other networks.

**Grad-CAM visualization results**. To have a clearer view of the attention region of networks, some images are visualized using the Grad-CAM [35]. The method provides a clear view of the area of interest of the networks, which demonstrates the capability of the networks to capture critical features. The results of visualization on the Kvasir dataset can be observed in Fig. 7. In contrast to the other three representative networks, the DFE-IANet can localize the critical location of the polyps with great precision. Besides, DFE-IANet can accurately localize the relatively weak features of ulcerative colitis. However, some other networks fail to identify them. Meanwhile, DFE-IANet can capture global features of z-lines, which shows the capability of global feature extraction. In contrast, although Swin-Transformer and CMT achieve global attention, these two networks are not sensitive to some critical features. These results show that the proposed DFE-IANet not only has the excellent capability of critical feature extraction but also has the advantages of local and global feature extraction.

## 5      Conclusion

In this study, a novel neural network based on dual-domain feature extraction and interaction attention, DFE-IANet, is proposed for polyp image classification. An innovative combination of spectral transformation and attention guidance aims to address the challenges of polyp image classification. The accuracies of DFE-IANet on the Kvasir dataset and CP dataset are 93.94% and 98.16%, respectively. DFE-IANet strikes the perfect balance between efficiency and accuracy, outperforming some classical and latest networks while reducing parameter numbers and computational costs. It is a better choice for precise polyp image classification in clinical applications. In the future, DFE-IANet will be extended to address other tasks in medical image analysis. This includes medical image segmentation, target detection, and localization, enabling the network to be utilized in a broader range of applications.

**Acknowledgments.** This work was supported National Science Innovation Special Zone Project (2019XXX00701); National Key Basic Research Program Project (624XXXX0206); Key research and development projects of Hunan Province (2020SK2134); Natural Science Foundation of Hunan Province (2019JJ80105, 2022JJ30625).

**Disclosure of Interests.** The authors have no competing interests to declare that are relevant to the content of this article.

**References**

1. Xi, Y., Xu, P.: Global colorectal cancer burden in 2020 and projections to 2040. TRANSL ONCOL **14**(10), 101174 (2021)
2. Pogorelov, K., Randel, K.R., Griwodz, C., Eskeland, S.L., de Lange, T., Johansen, D., Spampinato, C., Dang-Nguyen, D., Lux, M., Schmidt, P.T.: Kvasir: A multi-class image dataset for computer aided gastrointestinal disease detection. In: Proceedings of the 8th ACM on Multimedia Systems Conference, pp. 164-169 (2017)
3. Vaswani, A., Shazeer, N., Parmar, N., Uszkoreit, J., Jones, L., Gomez, A.N., Kaiser, A., Polosukhin, I.: Attention is all you need. Advances in neural information processing systems **30** (2017)
4. Patel, K., Li, K., Tao, K., Wang, Q., Bansal, A., Rastogi, A., Wang, G.: A comparative study on polyp classification using convolutional neural networks. PLOS ONE **15**(7), e0236452 (2020)
5. Wu, Y., Qi, S., Sun, Y., Xia, S., Yao, Y., Qian, W.: A vision transformer for emphysema classification using CT images. Physics in Medicine & Biology **66**(24), 245016 (2021)
6. Dosovitskiy, A., Beyer, L., Kolesnikov, A., Weissenborn, D., Zhai, X., Unterthiner, T., Dehghani, M., Minderer, M., Heigold, G., Gelly, S.: An image is worth 16x16 words: Transformers for image recognition at scale. arXiv preprint arXiv:2010.11929 (2020)
7. Komorowski, P., Baniecki, H., Biecek, P.: Towards evaluating explanations of vision transformers for medical imaging. In: Proceedings of the IEEE/CVF conference on computer vision and pattern recognition, pp. 3725-3731 (2023)
8. Liu, J., Li, Y., Cao, G., Liu, Y., Cao, W.: Feature pyramid vision transformer for medmnist classification decathlon. In: 2022 International Joint Conference on Neural Networks (IJCNN), pp. 1-8 (2022)
9. Wang, K., Zhuang, S., Ran, Q., Zhou, P., Hua, J., Zhou, G., He, X.: DLGNet: A dual-branch lesion-aware network with the supervised Gaussian Mixture model for colon lesions classification in colonoscopy images. MED IMAGE ANAL **87**, 102832 (2023)
10. Wei, J.W., Suriawinata, A.A., Vaickus, L.J., Ren, B., Liu, X., Lisovsky, M., Tomita, N., Abdollahi, B., Kim, A.S., Snover, D.C.: Evaluation of a deep neural network for automated classification of colorectal polyps on histopathologic slides. JAMA NETW OPEN **3**(4), e203398 (2020)
11. Wang, T., Lan, J., Han, Z., Hu, Z., Huang, Y., Deng, Y., Zhang, H., Wang, J., Chen, M., Jiang, H.: O-Net: a novel framework with deep fusion of CNN and transformer for simultaneous segmentation and classification. FRONT NEUROSCI-SWITZ **16**, 876065 (2022)

12. Huo, X., Sun, G., Tian, S., Wang, Y., Yu, L., Long, J., Zhang, W., Li, A.: HiFuse: Hierarchical multi-scale feature fusion network for medical image classification. BIOMED SIGNAL PROCES **87**, 105534 (2024)
13. Rippel, O., Snoek, J., Adams, R.P.: Spectral representations for convolutional networks. Advances in neural information processing systems **28** (2015)
14. Ding, X., Liu, Y., Zhao, J., Wang, R., Li, C., Luo, Q., Shen, C.: A novel wavelet-transform-based convolution classification network for cervical lymph node metastasis of papillary thyroid carcinoma in ultrasound images. COMPUT MED IMAG GRAP **109**, 102298 (2023)
15. Xu, K., Qin, M., Sun, F., Wang, Y., Chen, Y., Ren, F.: Learning in the frequency domain. In: Proceedings of the IEEE/CVF conference on computer vision and pattern recognition, pp. 1740-1749 (2020)
16. Huang, Z., Zhang, Z., Lan, C., Zha, Z., Lu, Y., Guo, B.: Adaptive Frequency Filters As Efficient Global Token Mixers. In: Proceedings of the IEEE/CVF International Conference on Computer Vision, pp. 6049-6059 (2023)
17. Yang, J., Li, C., Dai, X., Gao, J.: Focal modulation networks. Advances in Neural Information Processing Systems **35**, 4203-4217 (2022)
18. Rao, Y., Zhao, W., Tang, Y., Zhou, J., Lim, S.N., Lu, J.: Hornet: Efficient high-order spatial interactions with recursive gated convolutions. Advances in Neural Information Processing Systems **35**, 10353-10366 (2022)
19. Li, S., Wang, Z., Liu, Z., Tan, C., Lin, H., Wu, D., Chen, Z., Zheng, J., Li, S.Z.: MogaNet: Multi-order Gated Aggregation Network. In: The Twelfth International Conference on Learning Representations, (2023)
20. Dai, Z., Liu, H., Le, Q.V., Tan, M.: Coatnet: Marrying convolution and attention for all data sizes. Advances in neural information processing systems **34**, 3965-3977 (2021)
21. Yao, T., Pan, Y., Li, Y., Ngo, C., Mei, T.: Wave-vit: Unifying wavelet and transformers for visual representation learning. In: European Conference on Computer Vision, pp. 328-345 (2022)
22. Woo, S., Debnath, S., Hu, R., Chen, X., Liu, Z., Kweon, I.S., Xie, S.: Convnext v2: Co-designing and scaling convnets with masked autoencoders. In: Proceedings of the IEEE/CVF Conference on Computer Vision and Pattern Recognition, pp. 16133-16142 (2023)
23. Wang, P., Chen, P., Yuan, Y., Liu, D., Huang, Z., Hou, X., Cottrell, G.: Understanding convolution for semantic segmentation. In: 2018 IEEE winter conference on applications of computer vision (WACV), pp. 1451-1460 (2018)
24. Chu, X., Tian, Z., Zhang, B., Wang, X., Shen, C.: Conditional positional encodings for vision transformers. arXiv preprint arXiv:2102.10882 (2021)
25. Wang, W., Hu, Y., Luo, Y., Wang, X.: Medical image classification based on information interaction perception mechanism. COMPUT INTEL NEUROSC **2021** (2021)
26. Loshchilov, I., Hutter, F.: Decoupled weight decay regularization. arXiv preprint arXiv:1711.05101 (2017)
27. He, K., Zhang, X., Ren, S., Sun, J.: Deep residual learning for image recognition. In: Proceedings of the IEEE conference on computer vision and pattern recognition, pp. 770-778 (2016)
28. Liu, Z., Lin, Y., Cao, Y., Hu, H., Wei, Y., Zhang, Z., Lin, S., Guo, B.: Swin transformer: Hierarchical vision transformer using shifted windows. In: Proceedings of the IEEE/CVF international conference on computer vision, pp. 10012-10022 (2021)
29. Guo, J., Han, K., Wu, H., Tang, Y., Chen, X., Wang, Y., Xu, C.: Cmt: Convolutional neural networks meet vision transformers. In: Proceedings of the IEEE/CVF conference on computer vision and pattern recognition, pp. 12175-12185 (2022)
30. Wang, W., Dai, J., Chen, Z., Huang, Z., Li, Z., Zhu, X., Hu, X., Lu, T., Lu, L., Li, H.:

Internimage: Exploring large-scale vision foundation models with deformable convolutions. In: Proceedings of the IEEE/CVF Conference on Computer Vision and Pattern Recognition, pp. 14408-14419 (2023)
31. Ma, X., Zhou, Y., Wang, H., Qin, C., Sun, B., Liu, C., Fu, Y.: Image as set of points. arXiv preprint arXiv:2303.01494 (2023)
32. Yu, W., Zhou, P., Yan, S., Wang, X.: Inceptionnext: When inception meets convnext. arXiv preprint arXiv:2303.16900 (2023)
33. Zhu, L., Wang, X., Ke, Z., Zhang, W., Lau, R.W.: Biformer: Vision transformer with bi-level routing attention. In: Proceedings of the IEEE/CVF conference on computer vision and pattern recognition, pp. 10323-10333 (2023)
34. Liu, Y., Tian, Y., Zhao, Y., Yu, H., Xie, L., Wang, Y., Ye, Q., Liu, Y.: Vmamba: Visual state space model. arXiv preprint arXiv:2401.10166 (2024)
35. Selvaraju, R.R., Cogswell, M., Das, A., Vedantam, R., Parikh, D., Batra, D.: Grad-cam: Visual explanations from deep networks via gradient-based localization. In: Proceedings of the IEEE international conference on computer vision, pp. 618-626 (2017)